\documentclass{article}
\usepackage{PRIMEarxiv}
\usepackage[T1]{fontenc}
\usepackage{amsmath,amssymb,graphicx,booktabs,tabularx,longtable,array}
\usepackage{microtype}
\usepackage{etoolbox}
\usepackage{afterpage}
\usepackage{caption}
\usepackage[numbers,sort&compress]{natbib}
\usepackage[hidelinks]{hyperref}
\usepackage{xeCJK}
\xeCJKsetup{AutoFallBack=true}

\setCJKsansfont{bsmi00lp.ttf}
\setCJKmonofont{bsmi00lp.ttf}

\setCJKfallbackfamilyfont{\CJKrmdefault}
    {HaranoAjiMincho-Regular.otf}
\setCJKfallbackfamilyfont{\CJKsfdefault}
    {HaranoAjiMincho-Regular.otf}
\setCJKfallbackfamilyfont{\CJKttdefault}
    {HaranoAjiMincho-Regular.otf}

\DeclareMathSizes{10}{10}{7}{6}
\DeclareMathSizes{9}{9}{6}{6}
\DeclareMathSizes{8}{8}{6}{6}

\title{Contextual language representations link narrative processing to language and literacy in children}
\author{%
\normalfont Dongyi He\textsuperscript{\textdagger},
Tian Jiang\textsuperscript{\textdagger},
Nizhuan Wang\textsuperscript{*},
Wai Ting Siok\textsuperscript{*}\\
Department of Language Science and Technology, The Hong Kong Polytechnic University,\\
Hung Hom, Hong Kong SAR, China\\
\textsuperscript{\textdagger}\ These authors contributed equally to this work.\\
\textsuperscript{*}\ Corresponding authors:
\href{mailto:wangnizhuan1120@gmail.com}{wangnizhuan1120@gmail.com};
\href{mailto:wai-ting.siok@polyu.edu.hk}{wai-ting.siok@polyu.edu.hk}%
}
\date{}
\makeatletter

\begin{document}

\maketitle

\begin{abstract}
Spoken narratives extensively connect moment-to-moment language processing with knowledge accumulated through language and literacy experience. This study examined that relationship using functional magnetic resonance imaging (fMRI) during Cantonese story listening and contextual representations from large language model (LLM). The imaging cohort comprised 23 school-age children; 16 contributed complete standardized literacy and language assessments and served as held-out participants in a nested cross-subject encoding analysis of the same story. Contextual embeddings predicted distributed narrative responses, with prominent correspondence in temporoparietal and default-network parcels. Regional summaries of the top 25\% of voxels ranked by held-out correlation yielded atlas-wide means of $r=0.174$ and partial $R^2=0.0165$. A joint feature model identified negative surprisal and positive predictive-entropy coefficients across cortical and subcortical parcels, including control, salience, hippocampal and amygdalar regions. Higher word-dictation and overall literacy scores were associated with lower regional encoding performance. Higher Cantonese grammar scores were associated with more negative semantic-update coefficients in parietal control, somatomotor and dorsal-attention parcels. Together, these findings connect language and literacy skills to the regional strength and direction of neural coupling with narrative context. They position naturalistic context processing as a possibly measurable link between oral comprehension and literacy in childhood.

\end{abstract}


\section{Introduction}

Understanding a spoken story requires listeners to integrate successive words into representations of events, characters and goals \cite{willems2020narratives,baldassano2017event,hahamy2023reactivates,jamali2024semantic}. This integration draws on linguistic knowledge that develops alongside language and literacy skills during childhood \cite{skeide2016ontogeny,dehaene2015illiterate,malcorra2024speech}. A central question is how the neural representation of the unfolding spoken context relates to variation in language and literacy skills \cite{wat2024emergent,jangraw2023long}. Naturalistic narratives provide a setting in which shared stimulus-driven responses and individual differences can be examined within the same continuous experience \cite{simony2016dynamic,finn2018paranoia,finn2020idiosynchrony,song2026reading}. Story comprehension engages neural systems operating over multiple temporal scales. Narrated-story experiments reveal progressively longer temporal receptive windows from auditory to higher-order language and association cortex \cite{lerner2011topographic,goldstein2025temporal}. Continuous-speech encoding studies further show that semantic information is distributed across the cerebral cortex \cite{huth2016natural}. During narrative construction, information propagates across this hierarchy toward default-network regions that integrate extended context \cite{chang2022information}. Event-level representations in association cortex and hippocampal responses at event boundaries connect narrative understanding with memory formation \cite{baldassano2017event,park2025hippocampal}. These findings motivate a computational description that captures both the contextual state maintained across a story and its change as new information arrives \cite{tikochinski2025incremental,song2026reinstatement}.

Predictive models provide complementary descriptions of this evolving context. Surprisal quantifies the information carried by an observed linguistic unit relative to its preceding context, whereas entropy quantifies uncertainty over possible continuations \cite{hale2001surprisal,levy2008expectation}. Naturalistic fMRI studies associate these quantities with different regional response profiles \cite{willems2016prediction}. Electrophysiological evidence identifies predictions at nested linguistic levels, linking contextual expectations to phonological, syntactic and semantic processing \cite{heilbron2022hierarchy,zou2026constituent}. Related encoding work identifies a cortical hierarchy of predictive representations during speech comprehension \cite{caucheteux2023predictive,zhou2026hierarchical}. Together, these approaches motivate joint analysis of predictive uncertainty, observed information and representational updating. Contextual language models extend this approach by representing the preceding text as a high-dimensional state. Their internal representations capture combinations of lexical, syntactic and semantic information that can be aligned with continuous neural responses \cite{goldstein2025unified}. Previous studies demonstrate correspondence between language-model representations, fMRI responses and electrophysiological signals during language comprehension \cite{schrimpf2021neural,goldstein2022shared,caucheteux2022deep,gao2025alignment,gu2025reading}. Encoding models quantify how well contextual states predict neural responses, while feature-based models estimate signed associations with specified computational quantities. Combining these analyses connects the representation of narrative context with the neural response to its predictive properties.

Childhood provides a setting for examining how these computational descriptions relate to acquired language skills \cite{federmeier2026prediction}. Narrative fMRI studies identify coordinated temporal, frontal and parietal responses that change across the school years \cite{schmithorst2006narrative,szaflarski2012longitudinal}. Longitudinal evidence links early language-related and later higher-order network organization to reading proficiency \cite{horowitzkraus2024language}. The home reading environment is also associated with story-evoked responses in regions supporting semantic processing and mental imagery \cite{hutton2015home,horowitzkraus2026parental}. These findings connect naturalistic listening to the broader systems through which oral language and literacy develop \cite{besserilan2026storytelling}. 

Literacy experience and cultural specificity of a writing system can shape neural responses to spoken language. Studies comparing literacy levels identify changes in auditory language responses and recruitment of orthographic representations during speech processing \cite{dehaene2010literacy}. In children, reading experience is associated with differences in the cerebral organization of spoken-language processing \cite{monzalvo2013reading}. Chinese differs fundamentally from Indo-European alphabetic languages in its phonology, orthography, and grammar: it is a tonal language with a morphosyllabic writing system in which characters map onto syllables and morphemes rather than individual phonemes, and it employs an analytic, topic-prominent grammar that lacks inflectional morphology (such as tense or subject–verb agreement) and relies heavily on word order, aspect markers, and discourse context to convey syntactic and semantic relations. Neuroimaging studies using passive listening paradigms in Chinese, which are predominantly in Mandarin, have begun to characterize how these linguistic properties are represented in the brain. In fMRI, passive listening to spoken Mandarin sentences engages a predominantly ventral-stream temporal network encompassing the left middle temporal gyrus, left anterior temporal lobe, and bilateral posterior superior temporal regions, with relatively less dorsal inferior parietal involvement than typically reported in alphabetic languages \cite{Liu2020}. Extending to continuous narratives, the Le Petit Prince multilingual naturalistic fMRI corpus (LPPC-fMRI) established whole-brain neural tracking across acoustic, lexical, and syntactic levels during passive Mandarin audiobook listening in young adults \cite{Li2022}. Complementary electrophysiological work has demonstrated that cortical activity concurrently tracks hierarchical linguistic structures—syllables, phrases, and sentences—during connected Mandarin speech listening \cite{Ding2015}.

Within the broader Chinese language family, Cantonese, a major southern Chinese variety and the primary spoken language in Hong Kong, presents an even more complex suprasegmental and pragmatic profile: it features a rich nine-tone system alongside an extensive inventory of sentence-final particles that encode pragmatic and syntactic nuance. Despite this distinctiveness, neuroimaging research with passive listening paradigm in Cantonese remains sparse and has focused almost exclusively on adults. Passive oddball EEG studies have characterized pre-attentive mismatch negativity (MMN) responses to fine-grained Cantonese lexical tone contrasts (\cite{Zhang2018}), while sparse-sampling fMRI has examined suprasegmental phonetic feature representation during passive tonal syllable listening \cite{tao2026cantonese}. Most recently, the Le Petit Prince Hong Kong (LPPHK) dataset introduced multimodal naturalistic fMRI and electroencephalography (EEG) recordings from older Cantonese speakers passively listening to an audiobook, establishing neural alignment to continuous acoustic and lexical features in aging listeners (\cite{momenian2024petit}). However, naturalistic passive listening evidence in Cantonese-speaking children, particularly those linking continuous contextual representations to developing language and literacy skills, remains absent.

Three critical gaps remain in linking computational models of narrative processing to language and literacy development. First, neurocomputational studies using contextual language models and predictive features have focused overwhelmingly on adult listeners and alphabetic languages, leaving it largely unknown how these representational and predictive mechanisms operate in developing brains or in tonal, non-alphabetic language contexts such as Cantonese. Second, while developmental fMRI studies have demonstrated that narrative-evoked brain networks mature across childhood, few have applied fine-grained contextual encoding and joint predictive modelling directly to school-age child participants to test how individual neural coupling varies with acquired skills. Third, linking naturalistic brain responses to behavioral proficiency across a multi-year developmental span requires direct, norm-referenced standardized assessments. Unlike indirect environmental proxies or raw task scores that are confounded by chronological age, age-normed standardized batteries administered directly to children allow for fair, developmentally calibrated comparisons of oral language and literacy proficiency across age groups.

In the present study, we related Cantonese contextual narrative processing among school-age children to age-normed standardized measures of Chinese literacy and Cantonese spoken language. The present study combined story-listening fMRI from 23 children with representations from Gemma 3 1B \cite{gemma_2025}. A nested leave-one-subject-out analysis evaluated contextual encoding in 16 assessment-complete children. A joint voxelwise model estimated associations with surprisal, predictive entropy and semantic update, a geometric measure of contextual representational change. Regional brain--behavior analyses then related these measures to language, literacy and cognitive assessments. The design addressed how contextual representations map onto children's narrative responses and how regional model--brain relationships vary with language and literacy skills.

\section{Methods}

\begin{figure}[t]
\centering
\includegraphics[width=\linewidth]{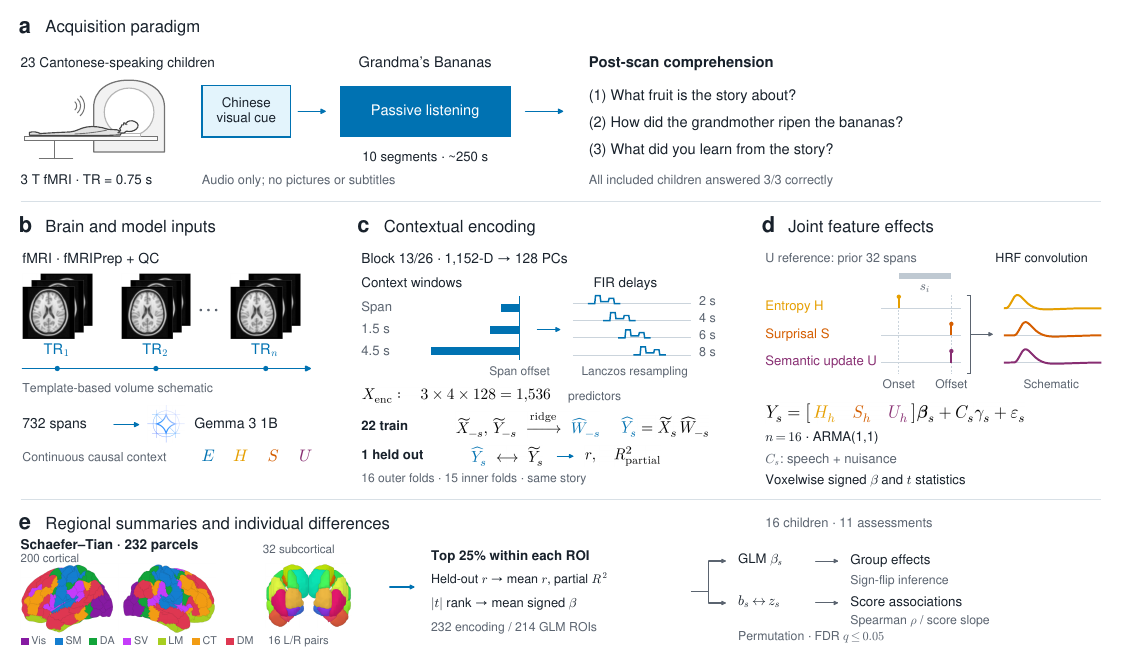}
\caption{\textbf{Naturalistic story-listening paradigm and analysis framework.} \textbf{a}, Cantonese-speaking children listened to a story during fMRI and completed post-scan comprehension questions. \textbf{b}, Story-aligned BOLD preprocessing and extraction of contextual embeddings and predictive features from Gemma 3 1B. MRI thumbnails schematically illustrate volume sampling using a public MNI152NLin6Asym anatomical template. \textbf{c}, Nested leave-one-subject-out encoding linked contextual embeddings to held-out responses; tildes denote nuisance-adjusted quantities. \textbf{d}, A joint voxelwise model estimated signed associations with surprisal, predictive entropy and semantic update. Single-span event curves illustrate haemodynamic convolution, and $\mathbf{C}_s$ denotes speech-related and nuisance regressors. \textbf{e}, Atlas-based regional summaries supported group-response and brain--behavior analyses using permutation inference and ROI-wise false discovery rate correction. Atlas colours identify cortical networks and subcortical subdivisions.}
\label{fig:fig1_workflow}
\end{figure}

This study integrated naturalistic story-listening fMRI, standardized behavioral assessments and computational language modelling to examine contextual language processing in Cantonese-speaking children (Figure~\ref{fig:fig1_workflow}). The imaging cohort comprised 23 children, including 16 who completed the full literacy, oral-language and cognitive assessment battery. Following preprocessing and quality assessment, blood-oxygenation-Level-dependent (BOLD) responses were aligned with contextual embeddings and predictive features extracted from Gemma 3 1B. Two complementary analyses characterized the relationship between these model-derived representations and neural responses. A joint voxelwise general linear model estimated signed associations with surprisal, predictive entropy and semantic update, conditional on the other features. Contextual-embedding encoding models used nested leave-one-subject-out cross-validation to evaluate prediction of the 16 assessment-complete children's responses to the same story. Both approaches accounted for speech-related, acoustic and nuisance variation. Selected-voxel summaries were then derived within cortical and subcortical regions defined by the Schaefer--Tian \cite{schaefer2018local,tian2020topographic} parcellation. With the child as the unit of inference, permutation analyses assessed regional text-feature responses and associations of regional encoding performance and text-feature coefficients with behavioral scores. False discovery rate (FDR) correction was applied across regions within each predefined family of tests.

\subsection{Data acquisition}

\subsubsection{Participants}
Participants were recruited from an ongoing cross-sectional study that included structural and functional MRI and formal assessments of general intelligence, literacy, and language development. All participants spoke Cantonese as their native, dominant, and family language, and attended local mainstream primary schools in Hong Kong where the curriculum was delivered in Cantonese. All children used English as a second language (reported mean age of acquisition [AoA] = 3.45 years) and learned Mandarin at school (reported mean AoA = 5.35 years). In addition, four of the 23 included participants reported learning experience with additional foreign languages, including French, Japanese, and Spanish. All participants reported normal hearing without any use of hearing aids. The study was approved by the Human Research Committee at The Hong Kong Polytechnic University (Reference Number: HSEARS20230814002), and written consent was obtained from each participant and their parent or legal guardian.

Five children were excluded for excessive head motion during scanning, and two were excluded because they fell asleep during scanning. The resulting story-imaging cohort comprised 23 children (12 girls and 11 boys, mean age = 10.59 years, SD = 1.56, range = 7.83–13.17 years).

Of these 23 children, 16 completed the full literacy and language assessments and constituted the brain–behavior cohort. Following the formal assessments, seven of the 16 children were diagnosed with specific learning disorder (SpLD), three of whom had comorbid developmental language disorder (DLD). These 16 children also served as held-out participants in the encoding-model evaluation, while the remaining seven contributed training data in every cross-validation fold (Supplementary Fig.~\ref{fig:cohort}). The brain–behavior cohort comprised eight girls and eight boys in grades 2–6 (mean age = 10.24 years, SD = 1.31, range = 8.42–12.25 years; Table~\ref{tab:sample}).

\begin{table}[htbp]\centering\small
\caption{Characteristics of the assessment cohort. All summaries refer to the 16 children used for brain--behavior analyses.}\label{tab:sample}
\begin{tabular}{lrr}\toprule Characteristic & Mean (SD) or count & Range \\\midrule
Age (years) & 10.24 (1.31) & 8.42--12.25 \\
Sex (female/male) & 8/8 & --- \\
School grade & 3.94 (1.39) & 2--6 \\\midrule
Full-scale IQ & 101.50 (11.62) & 85--128 \\
Literacy & 7.64 (3.26) & 1.75--13 \\
Word reading & 8.12 (3.77) & 1--15 \\
Word dictation & 6.31 (3.14) & 1--11 \\
Cantonese grammar & 11.75 (2.35) & 7--14 \\
Discourse comprehension & 11.31 (3.03) & 7--17 \\
Word definition & 12.12 (3.26) & 5--16 \\
Semantic relations & 8.50 (3.41) & 1--13 \\
Story retelling & 9.44 (3.14) & 1--14 \\
Noun expression & 9.88 (3.84) & 1--15 \\
HKGCNT & $-0.74$ (1.99) & $-5.07$--$1.57$ \\
\bottomrule\end{tabular}
\par\smallskip\begin{minipage}{.94\linewidth}\footnotesize Assessment values are the standardized scores supplied in the study assessment table. Age is derived from the recorded mean scan-month age.\end{minipage}
\end{table}

\subsubsection{MRI acquisition}
Whole-brain images were acquired on a 3.0-T Siemens Prisma scanner at The Hong Kong Polytechnic University. Functional data were acquired using an accelerated BOLD echo-planar imaging sequence (EPI; posterior-to-anterior phase encoding). The sequence comprised 60 interleaved slices with 2.4-mm isotropic voxels, a field of view of 216 × 216 mm, repetition time (TR) of 750 ms, echo time (TE) of 30 ms, and flip angle of 60°. A high-resolution T1-weighted structural image was acquired using a sagittal 3D MPRAGE sequence (240 slices; 0.8-mm isotropic voxels; field of view 256 × 224 mm; TR = 2300 ms; TE = 2.07 ms; inversion time = 1100 ms; flip angle = 7°; GRAPPA acceleration factor = 2).

\subsubsection{Story-listening paradigm}
Participants passively listened to a short story during fMRI scanning. A sound check was performed inside the scanner, during which the experimenter adjusted playback volume to each participant’s preferred comfortable level. A brief starting cue ("the story is about to start") was visually presented in Chinese on the screen before narration began.  The auditory-only presentation used Cantonese narration delivered in a soft female voice. The recording comprised ten audio segments lasting approximately 4 min 10 s in total.

The story, ``婆婆嘅香蕉'' (``Grandma's Bananas''), is classified as Level 4 in the Chinese Storybooks collection (https://global-asp.github.io/storybooks-chinese/), which assigns stories to five levels based on length and vocabulary. It follows a child narrator who discovers the grandmother’s secret banana-ripening method, takes bananas, and eventually learns not to steal bananas from the grandmother. The familiar family setting, concrete events, and relatively straightforward narrative sequence make the story appropriate for Cantonese-speaking pupils across Hong Kong Primary 2–6.

After scanning, all participants reported that this was their first exposure to the story. They were also asked three post-scan questions in Cantonese to assess story comprehension: (1) What fruit is the story about? (2) How did the grandmother ripen the bananas? and (3) What did you learn from the story? Only participants who answered all three comprehension questions correctly were included in the sample.

\subsubsection{Literacy, language, and cognitive assessment}
In Hong Kong, locally developed, standardized, norm-referenced assessment batteries are well established for evaluating both Chinese literacy \cite{ho2015hong} and Cantonese spoken language \cite{t2006hong} in school-age children. In clinical and educational practice, these instruments are routinely used to profile language and literacy development and to support the diagnosis of specific learning disorder and developmental language disorder. As they yield age-normed standard scores calibrated to local Hong Kong reference populations, these tools allow for direct, developmentally fair comparisons of individual language and literacy proficiency across primary school age groups.

Literacy was assessed using the Hong Kong Test of Specific Learning Difficulties in Reading and Writing for Primary School Students, Third Edition (HKT-P(III)). Chinese Word Reading and Chinese Word Dictation were the literacy tasks shared across the junior- and senior-primary batteries. These tasks and the overall General Literacy score were included in the analyses. Tables and figures label them word reading, word dictation and literacy, respectively. Scores are reported on the test's norm-referenced scale (mean = 10, SD = 3); scores of 7 or below were taken to indicate weakness. The HKT-P(III) has Hong Kong norms for children aged 6 years 1 month to 12 years 6 months \cite{ho2015hong}.

As an additional measure of reading, participants completed the Hong Kong Graded Character Naming Test (HKGCNT), which assesses oral reading of single Chinese characters of graded difficulty across primary grades 1–6 \cite{leung2008hong}. In this task, each child was asked to read aloud 150 single Chinese characters matching their grade level. Performance is reported as grade-referenced z-scores (mean = 0, SD = 1). In clinical practice and research in Hong Kong, a z-score of -1.5 or below is typically used as a benchmark to identify significant reading difficulties.

Cantonese oral-language ability was assessed using the Hong Kong Cantonese Oral Language Assessment Scale (HKCOLAS). The six subtests were Hong Kong Cantonese Grammar, Textual Comprehension, Word Definition, Lexical-Semantic Relations, Narrative, and Expressive Nominal Vocabulary. The corresponding analysis labels are Cantonese grammar, discourse comprehension, word definition, semantic relations, story retelling and noun expression. HKCOLAS is a norm-referenced assessment developed for Cantonese-speaking children in Hong Kong and standardized on 1,120 children. Subtest scores are reported on a scale with a mean of 10 and SD of 3; a score of 10 corresponds to the 50th percentile, and scores of 7 or below indicate weakness \cite{t2006hong}.

General intellectual ability was assessed using the Hong Kong version of the Wechsler Intelligence Scale for Children—Fourth Edition (WISC-IV(HK)). The WISC-IV assesses verbal comprehension, perceptual reasoning, working memory and processing speed, and provides a full-scale IQ index of general intellectual ability \cite{wechsler2010hong}. Full-scale IQ scores are standardized to a population mean of 100 (SD = 15), with scores of 85 or above indicating normal intellectual functioning.

Qualified practitioners administered the assessment and reported the results under the study's Hong Kong assessment protocol. Registered educational psychologist administered the HKT-P(III) and WISC-IV(HK), while registered speech therapist administered the HKCOLAS and HKGCNT. 

\subsection{MRI preprocessing and quality assessment}
Imaging data were organized according to the Brain Imaging Data Structure (BIDS) \cite{bids} and preprocessed using fMRIPrep version 25.2.7 \cite{fmriprep}. T1-weighted images underwent intensity non-uniformity correction with N4BiasFieldCorrection \cite{n4tustison2010} in ANTs 2.6.2 \cite{ants2011}, brain extraction using an OASIS template, and segmentation into grey matter, white matter and cerebrospinal fluid with FSL FAST \cite{fsl2012,fast2001}. FreeSurfer 7.3.2 \cite{freesurfer2012} was used for cortical surface reconstruction and refinement of the anatomical brain mask. Anatomical images were registered nonlinearly to standard space using ANTs.

For each functional run, fMRIPrep generated a BOLD reference image and estimated six rigid-body head-motion parameters with FSL MCFLIRT \cite{mcflirt2002}. Susceptibility distortion correction used the anatomically constrained, fieldmap-less SyN workflow \cite{syn2008}, with deformation restricted to the phase-encoding direction. The BOLD reference was aligned to the T1-weighted image using six-degree-of-freedom boundary-based registration \cite{bbr2009}. Slice timing was corrected with AFNI 3dTshift \cite{afni} to the beginning of each TR. Motion, distortion and spatial-registration transforms were composed for single-step volumetric resampling using cubic B-spline interpolation. The analyses used BOLD images in MNI152NLin6Asym space at 2-mm isotropic resolution.

Quality assessment included visual review of anatomical normalization, BOLD-to-anatomical alignment, coverage and artefacts in the fMRIPrep HTML reports. Quantitative checks assessed sequence length, framewise displacement (FD), standardized DVARS and temporal signal-to-noise ratio. Each selected story run contained 336 volumes at a TR of 0.75 s. Frames were flagged when FD exceeded 0.5 mm, standardized DVARS exceeded 1.5, or fMRIPrep identified a non-steady-state volume \cite{power}. The original acquisition-time grid was retained for stimulus alignment. Flagged frames were represented by individual spike regressors in the text-feature general linear model (GLM); encoding fitting and performance scoring used the unflagged rows of this same time grid.

Before statistical modelling, BOLD images were spatially smoothed with a 4-mm full-width-at-half-maximum Gaussian kernel normalized within each child's effective brain mask. Voxel time series were expressed as percent signal change (PSC), using the mean signal across unflagged frames as the reference. Common analysis masks were constructed by intersecting the effective masks of the 23 story runs. The encoding mask contained 224,757 voxels and the text-feature GLM mask contained 224,756 voxels.

\subsection{Language model and stimulus-aligned representations}
All reported analyses used Gemma 3 1B (\texttt{google/gemma-3-1b-pt}), the Schaefer--Tian parcellation and the top 25\% of voxels within each region od interest (ROI); encoding used 128 principal components (PC128). Model parameters and the fixed checkpoint revision are provided in Supplementary Table~\ref{tab:selected_parameters}. Feature extraction used the model's native tokenizer and continuous causal text context.

Character-level annotations provided timing for 869 characters. A common segmentation of 732 contiguous text spans linked these annotations to model features. Each span inherited the onset of its first character and the offset of its final character. Contextual embeddings were extracted at the final token of the exact text prefix ending at each span, from decoder block 13 of 26. Each embedding contained 1,152 dimensions. Prefixes crossing native tokenization boundaries were separately tokenized and evaluated at their own boundaries. This procedure aligned representations with the annotated spans while retaining native tokenization.

Three scalar features described the story's model-derived predictive and contextual dynamics. For span $i$, surprisal was the sum of native token surprisals assigned to that span,
\begin{equation}
 S_i=-\sum_{k\in\mathcal{T}_i}\log_2 P(w_k\mid w_{<k}),
\end{equation}
where $\mathcal{T}_i$ contains tokens whose final character falls within span $i$. Each token contributed once, at the offset of the corresponding common span. Predictive entropy quantified the uncertainty of the native next-token distribution conditional on the text preceding span onset,
\begin{equation}
 H_i=-\sum_{w\in\mathcal{V}}P(w\mid c_i)\log_2 P(w\mid c_i),
\end{equation}
where $\mathcal{V}$ is the model vocabulary and $c_i$ is the exact preceding text prefix.

Semantic update quantified the change in the direction of a recent contextual representation after incorporating the current span. Let $\mathbf{u}_i$ denote the L2-normalized contextual embedding. The preceding context and its update were defined as
\begin{equation}
 \mathbf{m}_i=\frac{1}{32}\sum_{j=i-32}^{i-1}\mathbf{u}_j,
 \qquad
 \mathbf{m}_i^{+}=\frac{32\mathbf{m}_i+\mathbf{u}_i}{33},
 \qquad
 U_i=1-\frac{\mathbf{m}_i^{\mathsf{T}}\mathbf{m}_i^{+}}
 {\lVert\mathbf{m}_i\rVert\,\lVert\mathbf{m}_i^{+}\rVert}.
\end{equation}
Semantic update was evaluated from span 33 onward, providing a complete preceding 32-span context for each analysed event.

\subsection{Voxelwise text-feature GLM}
For each child, surprisal, entropy and semantic update were entered jointly into a voxelwise GLM. Feature amplitudes were standardized across spans 33--732. Entropy events were placed at span onsets; surprisal and semantic-update events were placed at span offsets. Event trains were convolved with a peak-normalized canonical double-gamma haemodynamic response function (HRF), comprising gamma components with shape parameters 6 and 16 and a 6:1 peak-to-undershoot weighting. The text-feature GLM used 316 BOLD volumes beginning at 15.0 s, the first sampled time point at or after the onset of span 33.

The model included four speech-related control regressors: span-event density, span duration, span token count and the audio root-mean-square envelope. The envelope was calculated in 20-ms windows from audio sampled at 16 kHz. Speech control regressors were HRF-convolved. Motion regressors comprised six parameters, their first derivatives, and the squares of both sets, yielding 24 motion terms. Additional nuisance terms included mean white-matter and cerebrospinal-fluid signals, discrete cosine drift regressors with a 128-s high-pass period, frame-specific spike regressors and an intercept. Thus, each text-feature coefficient estimated the response associated with that feature conditional on the other text features and nuisance terms.

Models were fitted with AFNI 3dREMLfit using voxelwise autoregressive moving-average ARMA(1,1) temporal-noise modelling and generalized least squares\cite{afni,arma}. The outputs included signed PSC coefficients and corresponding $t$ statistics for each of the three text features. The 16 first-level fits yielded 48 feature-specific coefficient maps.

\subsection{Contextual-embedding encoding models}
Encoding models predicted voxelwise BOLD responses from the first 128 principal components of the Gemma 3 1B contextual embeddings. Principal component analysis was fitted to the centred 732-span embedding matrix of the story; these components retained 91.97\% of its variance (Supplementary Table~\ref{tab:selected_parameters}).

To incorporate multiple scales of preceding context, three embedding series were constructed: the current span embedding and the mean embeddings within trailing 1.5-s and 4.5-s windows ending at each span offset. Each series was resampled to fMRI time points with a three-lobe Lanczos filter and expanded with finite impulse response (FIR) delays of 2, 4, 6 and 8 s. For span offset $o_i$, BOLD sampling time $t$ and delay $\ell$, the interpolation weight was
\begin{equation}
 L(u)=\operatorname{sinc}(u)\operatorname{sinc}(u/3)\,
       \mathbb{I}(|u|<3)\mathbb{I}(o_i\leq t),
 \qquad u=\frac{t-\ell-o_i}{\mathrm{TR}},
\end{equation}
with $\operatorname{sinc}(u)=\sin(\pi u)/(\pi u)$. Combining the three context scales and four delays produced 12 feature blocks and 1,536 expanded columns. Each expanded column was centred and scaled across the 324 retained sampling times, beginning at 9.0 s and ending at 251.25 s. An exact singular-value reparameterization expressed the design in its nonzero temporal subspace while preserving the expanded feature kernel and ridge penalty.

BOLD and stimulus predictors were projected into each child's nuisance-residual space using the four speech controls, 24 motion terms, white-matter and cerebrospinal-fluid signals, cosine drifts and intercept. The projection was estimated on that child's unflagged frames. Voxelwise ridge coefficients minimized
\begin{equation}
 \widehat{\boldsymbol{\beta}}_v
 =\operatorname*{arg\,min}_{\boldsymbol{\beta}}
 \left\{\frac{1}{N_{\mathrm{train}}}
 \sum_{s\in\mathcal{S}_{\mathrm{train}}}
 \bigl\lVert\widetilde{\mathbf{y}}_{sv}-\widetilde{\mathbf{X}}_s\boldsymbol{\beta}\bigr\rVert^2
 +\alpha\lVert\boldsymbol{\beta}\rVert^2\right\},
\end{equation}
where $N_{\mathrm{train}}$ is the total number of unflagged training frames and tildes denote nuisance-adjusted quantities. Candidate regularization strengths were $\{10^{-3},10^{-2},\ldots,10^{4}\}$.

Nested leave-one-subject-out (LOSO) cross-validation produced predictions for each of the 16 children with complete assessments. Each outer fold trained on the other 22 children. Regularization was selected separately for each voxel by minimizing summed squared error across 15 inner validation folds. Each inner fold withheld one remaining child with complete assessments. The seven additional children contributed to every training fold. Models were then refitted on all 22 outer-fold training children and evaluated on the held-out child's response to the same story.

Two metrics summarized each held-out voxel. Pearson's $r$ measured correspondence between predicted and measured nuisance-adjusted BOLD. Incremental prediction relative to the nuisance baseline was expressed as cross-validated partial $R^2$,
\begin{equation}
 R^2_{\mathrm{partial},v}
 =1-\frac{\sum_{t\in\mathcal{G}_s}(\widetilde y_{svt}-\widehat{\widetilde y}_{svt})^2}
 {\sum_{t\in\mathcal{G}_s}\widetilde y_{svt}^{\,2}}
 =\frac{\mathrm{SSE}_{\mathrm{nuisance}}-\mathrm{SSE}_{\mathrm{full}}}
 {\mathrm{SSE}_{\mathrm{nuisance}}},
\end{equation}
where $\mathcal{G}_s$ denotes the held-out child's unflagged frames. Both measures were carried forward at their observed signed values.

\subsection{Whole-brain parcellation and regional summaries}
Regional analyses used Schaefer's 200-region cortical atlas combined with the 32-region Tian scale-2 subcortical atlas \cite{schaefer,tian}. Atlas labels were aligned with the common 2-mm analysis grid. Encoding analyses included all 232 represented regions. Text-feature GLM analyses required at least 80\% of a region's atlas voxels to fall within the common mask, yielding 214 eligible regions.

Within each region of interest (ROI), the top 25\% of covered voxels were summarized. The selected count was the ceiling of 0.25 multiplied by the number of covered ROI voxels. For text-feature GLMs, voxels were ranked separately for each child and feature by the absolute first-level $t$ statistic; the signed coefficients of the selected voxels were then averaged. For encoding models, voxels were ranked by that child's held-out Pearson correlation, and the selected voxels' $r$ and partial $R^2$ values were averaged arithmetically. Voxel selection used each child's imaging statistic; assessment scores entered the subsequent association analysis.

Encoding performance maps show the arithmetic mean of each regional summary across the 16 held-out children. Atlas-wide descriptive summaries gave equal weight to the 232 ROIs within each child. The mean and sample SD were then calculated across children. Source data provide each regional mean and SD.

Regional values were displayed on the inflated fsaverage cortical surface (10,242 vertices per hemisphere). MNI152 atlas labels were projected using registration-fusion coordinates distributed with neuromaps \cite{neuromaps}. Nearest-neighbour sampling preserved discrete parcel identities during projection onto the surface. Values were assigned to the corresponding surface parcels and the medial wall was masked. Left and right lateral and medial views used fixed camera positions. Text-feature group maps display BH-FDR-significant cortical results with direct ROI names and a common signed $t$ scale; subcortical results appear in named statistic matrices. Brain--behavior location maps identify the ROI and association direction, with adjacent individual-level scatterplots and effect summaries. Scatterplots include descriptive ordinary least-squares lines with pointwise 95\% Student's $t$ confidence bands for the fitted mean. Marginal Gaussian kernel-density estimates use Scott bandwidths and are displayed over the observed value ranges, with rug marks for individual observations. All regional estimation and inference used the original volumetric analysis grid.

\subsection{Regional inference and multiple-comparison correction}
The child was the unit of inference. Text-feature group responses were assessed with an intercept-only second-level model of the 16 children's regional coefficients. Two-sided permutation probabilities were obtained from all 32,768 distinct sign-pattern representatives after accounting for equivalence of opposite signs. Assessment associations were estimated separately for each score using
\begin{equation}
 \overline{\beta}_{sr}=a_r+b_r z_s+\epsilon_{sr},
\end{equation}
where $\overline{\beta}_{sr}$ is the child's selected-voxel regional coefficient and $z_s$ is the assessment score centred and scaled by its sample SD. The slope $b_r$ expresses the change in regional coefficient per SD of assessment score. Two-sided score-permutation probabilities used 19,999 random permutations plus the observed assignment. Analytic 95\% confidence intervals for group means and slopes were computed with the corresponding Student's $t$ distribution.

Encoding--assessment associations were quantified with Spearman's rank correlation across the 16 children. Exact two-sided probabilities were computed from the full label-permutation distribution using a dynamic-programming calculation of rank-dot-product counts. Tied ranks were assigned their average values, and the permutation distribution retained the multiplicity of individual label assignments.

The Benjamini--Hochberg false discovery rate (BH-FDR) procedure \cite{bh} was applied at $q\leq0.05$. Each text-GLM family comprised the 214 eligible ROIs for one text feature and one group or individual-assessment contrast. Each encoding family comprised the 232 ROIs for one performance metric and one assessment. Group-response and brain--behavior plots display results passing these ROI-wise thresholds. Each significant assessment association is accompanied by the corresponding individual participant values. Encoding-performance panels display descriptive cross-validated estimates. Supplementary Table~\ref{tab:counts} reports the complete test totals for this specification; source data provide regional estimates and probabilities.

\section{Results}

The story-imaging cohort comprised 23 children. Sixteen children completed all 11 literacy, oral-language and cognitive assessments (Table~\ref{tab:sample}). All reported analyses used Gemma 3 1B, the Schaefer--Tian atlas, the top 25\% of voxels within each ROI and PC128 embeddings for encoding. Cohort distributions are shown in Supplementary Fig.~\ref{fig:cohort}.

\subsection{Contextual embeddings and predictive feature profiles across the narrative}

\begin{figure}[t]
\centering
\includegraphics[width=\linewidth]{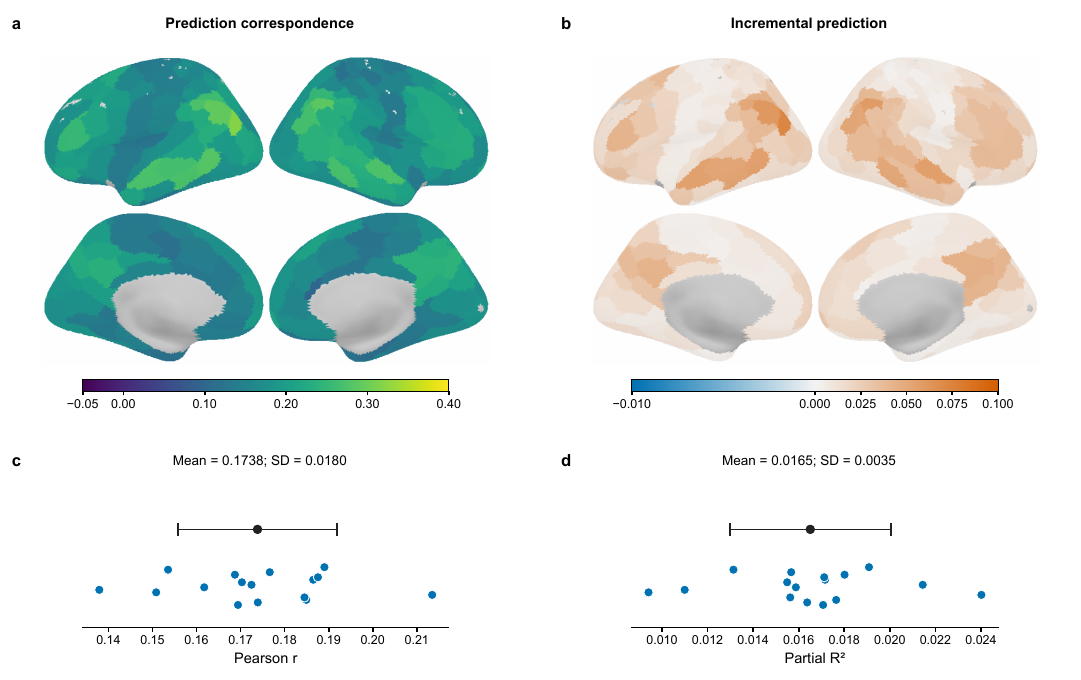}
\caption{\textbf{Contextual embeddings captured spatially structured story responses.}
\textbf{a,b}, Mean regional Pearson $r$ and partial $R^2$ for Gemma 3 1B/PC128 embeddings, summarized from the top 25\% of covered voxels within each Schaefer--Tian parcel. Maps show cortical parcels on inflated fsaverage surfaces, with left/right hemispheres in columns and lateral/medial views in rows. Both metrics summarize the same voxels, ranked separately for each held-out child by prediction $r$. \textbf{c,d}, Individual atlas-wide summaries, calculated by averaging equally across all 232 represented cortical and subcortical ROIs within each child. Blue points show the 16 children; black points and bars show the mean and one SD. These displays describe selected-voxel prediction performance.}
\label{fig:fig2_encoding}
\end{figure}

Contextual embeddings yielded spatially structured prediction of held-out story responses, with prominent correspondence in temporal and parietal parcels (Fig.~\ref{fig:fig2_encoding}). Averaged equally across the 232 ROIs within each child, prediction $r$ was $0.174\pm0.018$ and partial $R^2$ was $0.0165\pm0.0035$ (mean $\pm$ SD across 16 held-out children). Both metrics summarized the same voxels, selected within each ROI by held-out prediction $r$.

The left default-network parietal parcel (\texttt{LH\_Default\_Par\_2}) showed the largest regional correspondence, with mean $r=0.318$ and partial $R^2=0.0718$. Additional left and right default-network parietal parcels also showed prominent performance. The positive partial $R^2$ values quantify incremental prediction relative to the nuisance baseline.

\begin{figure}[t]
\centering
\includegraphics[width=\linewidth]{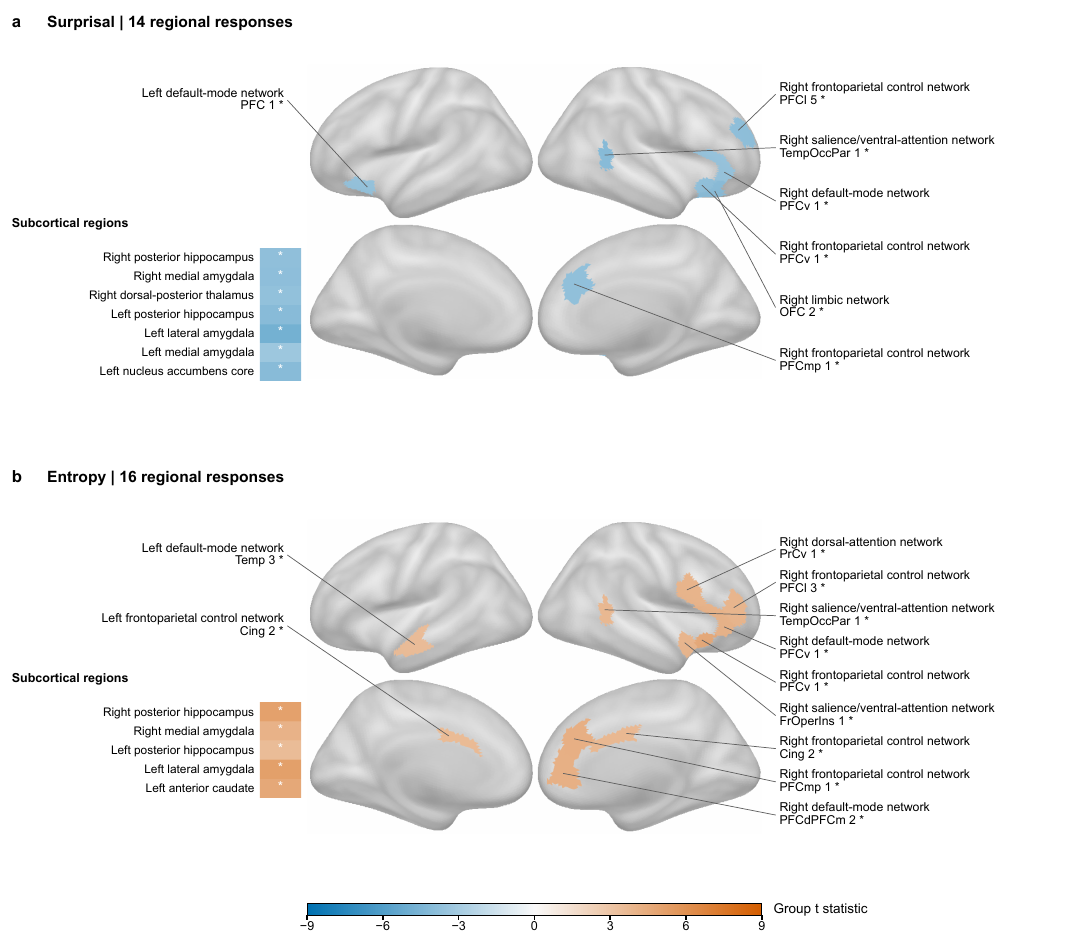}
\caption{\textbf{Surprisal and entropy mapped signed regional response profiles.}
\textbf{a,b}, Complete group responses passing ROI-wise BH-FDR for surprisal and entropy in the joint three-feature GLM ($n=16$). All panels summarize the top 25\% of covered voxels in each Schaefer--Tian parcel using Gemma 3 1B features. Colours show signed group $t$ statistics on a shared scale. Cortical parcels appear on inflated standard surfaces with direct network-parcel names and leaders; named matrices show subcortical results. Maps show left/right hemispheres in columns and lateral/medial views in rows. Two-sided probabilities used exact sign flips; BH adjustment included all 214 eligible ROIs separately for each feature. Stars denote the resulting $q$ values: $^{*}q\leq0.05$, $^{**}q\leq0.01$, $^{***}q\leq0.001$. Supplementary Table~\ref{tab:selected_group} lists all 30 displayed coefficients, confidence intervals and probabilities.}
\label{fig:fig3_text}
\end{figure}

The joint three-feature GLM identified 30 significant regional responses after BH-FDR correction within feature-specific ROI families (642 tests overall) (Fig.~\ref{fig:fig3_text}; Supplementary Table~\ref{tab:counts}). These responses comprised 14 negative surprisal coefficients and 16 positive entropy coefficients. All signed effects were estimated conditional on the other text features and nuisance regressors.

In the right ventral prefrontal control-network parcel, surprisal had a negative coefficient ($\overline{\beta}=-0.0353$, $q=0.0448$), whereas entropy had a positive coefficient ($\overline{\beta}=0.0275$, $q=0.0327$). The right temporo-occipito-parietal salience/ventral-attention parcel showed the same directional profile, with $\overline{\beta}=-0.0382$ for surprisal ($q=0.0448$) and $0.0266$ for entropy ($q=0.0418$). Additional significant responses occurred in bilateral posterior hippocampal and amygdalar subdivisions.

Encoding correspondence was prominent in temporoparietal and default-network parcels. Significant feature coefficients extended across control, attention, default-network and subcortical parcels.

\subsection{Literacy and grammar were associated with model--brain measures}

\begin{figure}[t]
\centering
\includegraphics[width=\linewidth]{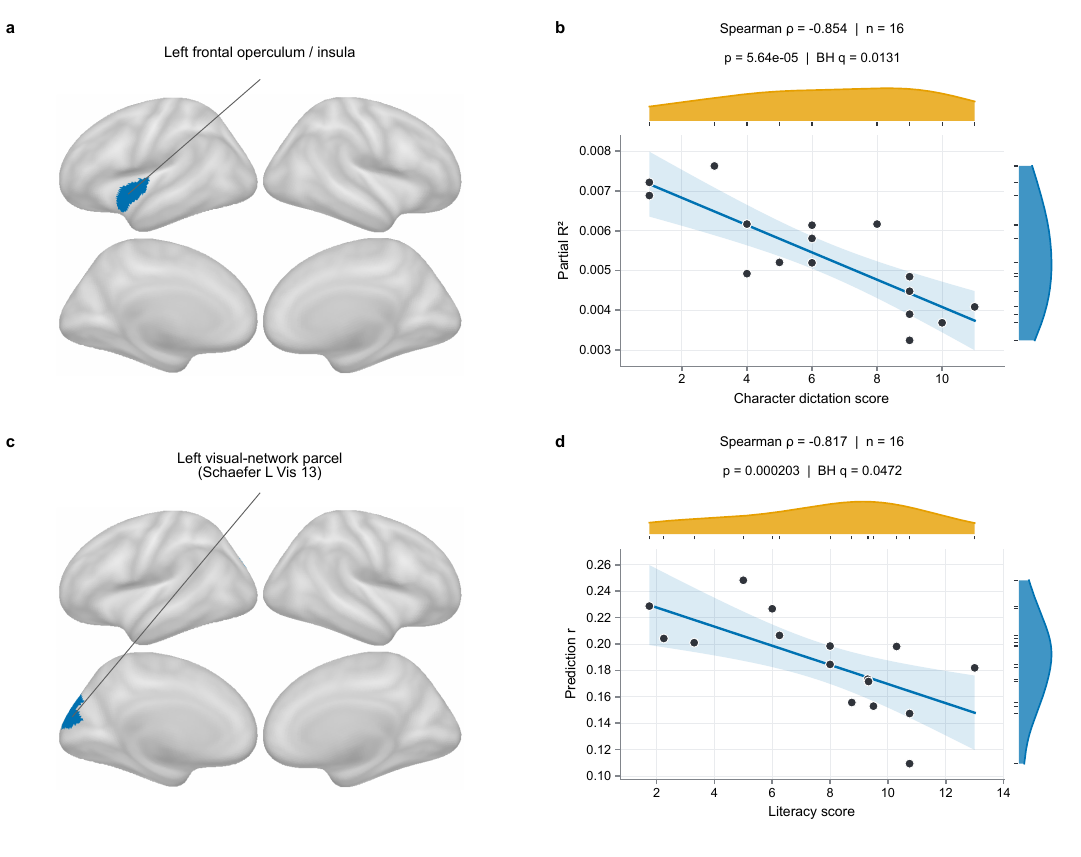}
\caption{\textbf{Regional encoding performance was associated with literacy scores.}
\textbf{a,c}, Named cortical locations of two significant encoding--assessment associations. \textbf{b}, Left frontal operculum/insula partial $R^2$ versus character dictation. \textbf{d}, Left visual-network prediction $r$ versus literacy. Each point represents one child ($n=16$). Blue lines and shaded bands show descriptive ordinary least-squares fits and pointwise 95\% confidence intervals for the fitted mean. Upper orange and right blue marginals show Gaussian kernel-density estimates with Scott bandwidths; rug marks indicate individual observations. Statistical annotations report Spearman correlations and exact two-sided label-permutation probabilities. Displayed $q$ values reflect BH-FDR correction across 232 ROIs for each metric--assessment pair. All panels use Gemma 3 1B/PC128, Schaefer--Tian and top 25\%. Supplementary Table~\ref{tab:associations} gives the exact raw and adjusted probabilities.}
\label{fig:fig4_encoding_behavior}
\end{figure}

Encoding--assessment analyses identified two significant associations after BH-FDR correction within metric--assessment ROI families (5,104 tests overall) (Fig.~\ref{fig:fig4_encoding_behavior}; Supplementary Table~\ref{tab:counts}). Character dictation was negatively associated with partial $R^2$ in the left frontal operculum/insula parcel (Spearman $\rho=-0.854$, $q=0.0131$). Literacy was negatively associated with prediction $r$ in a left visual-network parcel ($\rho=-0.817$, $q=0.0472$). The individual-level plots locate these associations in fronto-insular and visual-network parcels, and Supplementary Table~\ref{tab:associations} reports the corresponding raw and adjusted probabilities.

\begin{figure}[t]
\centering
\includegraphics[width=.95\linewidth]{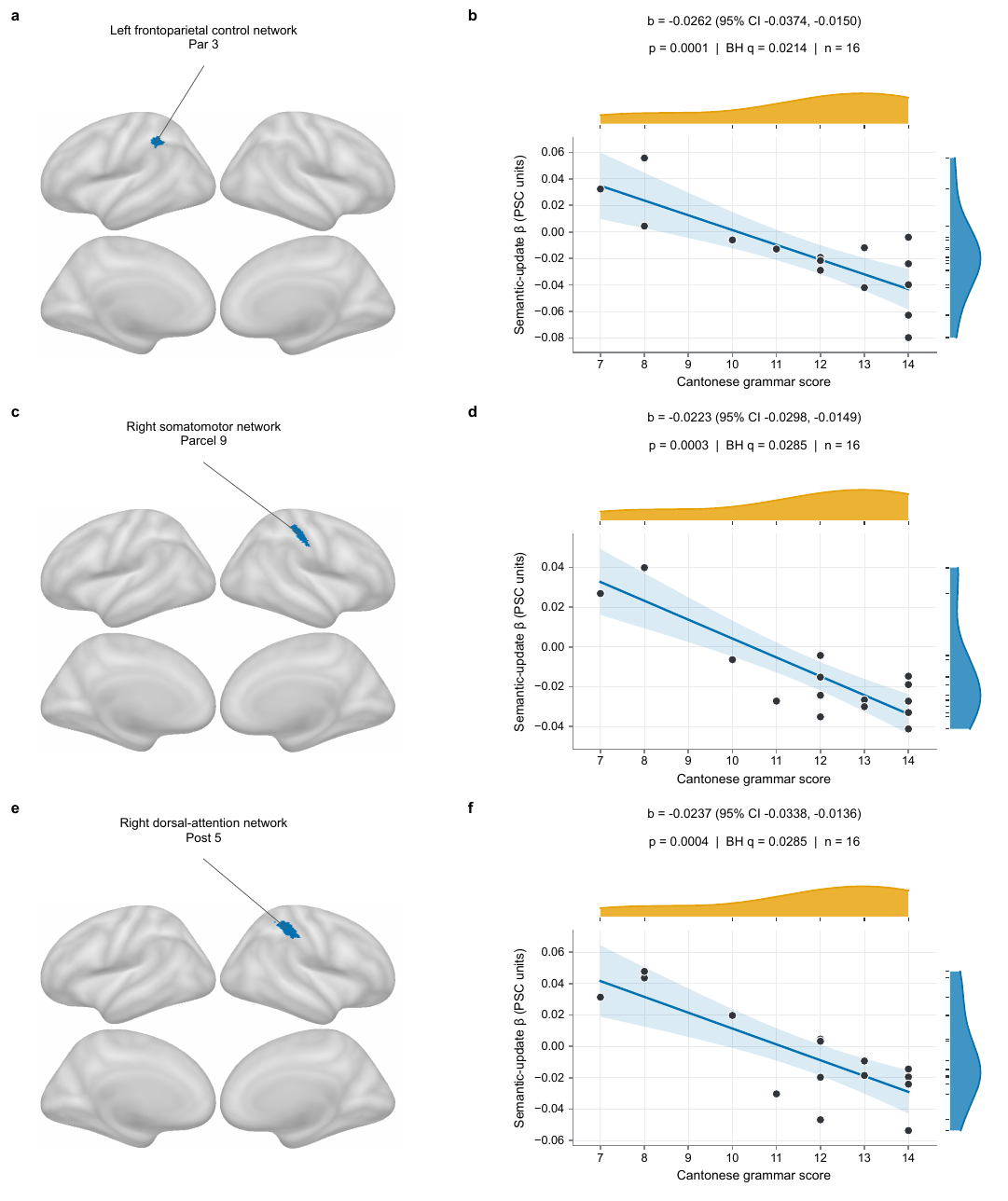}
\caption{\textbf{Semantic-update coefficients were associated with grammar scores.}
\textbf{a,c,e}, Named parcels in the left parietal control, right somatomotor and right posterior dorsal-attention networks. Blue indicates the negative association direction. \textbf{b,d,f}, Individual regional coefficients versus grammar scores for the same 16 children, using Gemma 3 1B, Schaefer--Tian and top 25\%. Lines and shaded bands show ordinary least-squares fits and pointwise 95\% confidence intervals for the fitted mean. Upper orange and right blue marginals show Gaussian kernel-density estimates with Scott bandwidths; rug marks indicate individual observations. Slopes $b$ express changes in the signed semantic-update coefficient per sample SD of grammar score. Annotated slope intervals are analytic 95\% Student's $t$ intervals (14 degrees of freedom). Two-sided probabilities were obtained by score permutations across the assessment cohort. BH-FDR correction included all 214 eligible ROIs for this feature--assessment pair. Higher scores correspond to more negative coefficients in all three parcels. Supplementary Table~\ref{tab:associations} gives the original raw and adjusted probabilities.}
\label{fig:fig5_semantic_behavior}
\end{figure}

Text-GLM assessment analyses identified three significant associations after BH-FDR correction within feature--assessment ROI families (7,062 tests overall) (Fig.~\ref{fig:fig5_semantic_behavior}). All three linked semantic-update coefficients to Cantonese grammar scores. The left parietal control-network parcel had a slope of $b=-0.0262$ PSC coefficient units per sample SD of grammar score (95\% CI, $-0.0374$ to $-0.0150$; $q=0.0214$). Corresponding slopes were $-0.0223$ in a right somatomotor parcel (95\% CI, $-0.0298$ to $-0.0149$; $q=0.0285$) and $-0.0237$ in a right posterior dorsal-attention parcel (95\% CI, $-0.0338$ to $-0.0136$; $q=0.0285$). Higher grammar scores were associated with more negative semantic-update coefficients across all three parcels.

\section{Discussion}

By combining contextual language modelling with direct, age-normed standardized assessments in school-age Cantonese-speaking children, these findings extend neurocomputational accounts of naturalistic language processing beyond adult alphabetic cohorts and demonstrate how individual differences in oral language and literacy map onto neural coupling with an unfolding narrative.

The results connect language and literacy skills with two properties of naturalistic neural responses: correspondence with a shared contextual representation and signed modulation by contextual change. These relationships occurred within distributed story responses and complementary entropy--surprisal profiles. Together, they link behavioral proficiency with how different regions respond to an unfolding narrative.

\subsection{Contextual representations connect shared narrative responses with extended integration}

The encoding results identify a component of narrative response that generalizes across listeners exposed to the same unfolding story. The strongest correspondence occurred in a left default-network parietal parcel, with selected-voxel mean $r=0.318$ and partial $R^2=0.0718$. Additional temporoparietal parcels showed prominent correspondence after adjustment for speech timing, acoustic energy and nuisance variation. This distribution connects model-derived contextual information with association systems involved in integrating meaning over extended intervals.

The anatomical pattern is consistent with converging accounts of narrative integration. Temporal receptive-window studies place association cortex toward the longer-timescale end of the cortical hierarchy \cite{lerner2011topographic}. Narrative-construction analyses further relate this organization to information flow from auditory and language regions toward the default network \cite{chang2022information}. Event-segmentation research identifies relatively stable event representations in higher-order cortex and hippocampal responses around transitions between events \cite{baldassano2017event}. Against this background, prominent contextual correspondence in default-network parietal cortex is consistent with a role for maintaining information across successive narrative units. Contextual embeddings provide a continuous representational description of that integration.

\afterpage{\clearpage}

The cross-subject design adds a developmental and cross-linguistic perspective to established language-model encoding approaches. Whereas prior contextual encoding studies have focused largely on adult listeners in alphabetic languages \cite{schrimpf2021neural},\cite{goldstein2022shared},\cite{caucheteux2022deep}, recent naturalistic work in Chinese has begun to characterize narrative tracking in young Mandarin-speaking adults \cite{Li2022} and older Cantonese-speaking adults \cite{momenian2024petit}, alongside sentence-level fMRI evidence of ventral-stream temporal engagement during passive Mandarin listening \cite{Liu2020}. The present study extends this line of inquiry to school-age Cantonese listeners, demonstrating that a shared contextual representation predicts distributed temporoparietal and default-network story responses in developing brains.

\subsection{Predictive variables characterize complementary regional response profiles}

The joint feature analysis distinguishes three model-derived properties of narrative context. Entropy describes uncertainty over the next-token distribution, surprisal describes the information carried by an observed span, and semantic update describes change in a contextual representation. Their joint estimation assigns each coefficient a conditional interpretation relative to the other predictors. The observed positive entropy and negative surprisal coefficients therefore describe how regional responses covaried with different aspects of the same unfolding narrative.

The distinction between uncertainty and observed information provides the clearest interpretation of the signed profiles. In the right ventral prefrontal control and temporo-occipito-parietal salience/ventral-attention parcels, BOLD responses increased with entropy and decreased with surprisal, conditional on the model terms. Greater uncertainty is compatible with greater engagement of systems involved in selecting among possible continuations. The surprisal coefficients characterize a separate conditional relationship with the information supplied by the observed span. Earlier naturalistic work established the utility of analysing entropy and surprisal separately \cite{willems2016prediction}. Electrophysiological and encoding studies additionally place prediction within a hierarchy spanning linguistic levels and cortical timescales \cite{heilbron2022hierarchy,caucheteux2023predictive}. The present findings extend this computational differentiation to children's narrative responses and emphasize the value of retaining signed regional estimates.

Hippocampal and amygdalar findings broaden the relevant systems beyond cortical language processing. Event-segmentation research links hippocampal activity at narrative transitions to the encoding of preceding events \cite{baldassano2017event}. This provides a basis for interpreting hippocampal predictive-feature associations in relation to the interaction between contextual expectations and event memory. Amygdalar sensitivity to surprisal has also been reported during natural story listening \cite{willems2016prediction}. The present subcortical results place uncertainty-related response variation within this wider narrative-processing system. A candidate account is that predictive context influences both ongoing interpretation and the prioritization of information for memory. This account generates a testable connection between linguistic prediction and the organization of remembered events.

\subsection{Literacy-related variation concerns the organization of model--brain correspondence}

The literacy findings establish a relationship between regional contextual encoding and behavioral proficiency during auditory story comprehension. Higher dictation scores were associated with lower partial $R^2$ in the left frontal operculum/insula, and higher literacy scores with lower prediction $r$ in a left visual-network parcel. These associations describe variation in the correspondence between a shared model-derived representation and each child's neural response. Their direction makes the regional and computational definition of model--brain correspondence central to its behavioral interpretation.

Evidence for interactions between literacy and spoken language gives these associations a broader context. Literacy acquisition is associated with changes in auditory language responses and access to orthographic representations during spoken input \cite{dehaene2010literacy}. Children's reading experience also relates to the organization of spoken-language responses \cite{monzalvo2013reading}. Chinese studies complement this evidence by linking writing skill to reading performance and identifying distributed neural systems engaged during character production \cite{tan2005writing,yang2019writing}. These findings support considering oral narrative processing within a system shaped by both spoken-language and literacy experience. In contrast to alphabetic scripts where literacy relies primarily on phoneme–grapheme mapping, Chinese employs a morphosyllabic writing system in which character acquisition depends heavily on visuospatial analysis and sensorimotor writing practice \cite{tan2005writing}\cite{yang2019writing}. The involvement of a left visual-network parcel with overall literacy and a left fronto-insular parcel with character dictation is consistent with this orthographic profile, suggesting that individual differences in Chinese character-writing and reading proficiency modulate how visual-orthographic and articulatory–motor systems couple with spoken narrative context.

The negative associations are informative in light of the cross-subject encoding design. Predictions reflect a mapping learned from other listeners, and higher literacy scores corresponded to lower agreement with that mapping in the identified parcels. One candidate explanation is a shift in the contribution of listener-specific phonological, articulatory or visually mediated representations during listening. This interpretation connects the association direction to how literacy-related knowledge is incorporated into a shared narrative. The fronto-insular and visual-network locations motivate testing that account within a distributed language system.

\subsection{Grammar proficiency relates to signed coupling with contextual change}

The semantic-update findings connect grammar proficiency with response modulation by changes in a model-defined contextual representation. Higher grammar scores corresponded to more negative coefficients in left parietal control, right somatomotor and right dorsal-attention parcels. This common slope direction across three systems identifies a distributed relationship between grammatical skill and contextual change. It complements the literacy associations by relating behavioral variation to a dynamic property of the representation.

Semantic update measures the directional change of a recent contextual average after a span is incorporated. This definition makes it sensitive to the relationship between new information and the accumulated context. Grammatical knowledge offers constraints on how successive elements can be combined into structured interpretations. Neural studies of constituent construction show that linguistic responses vary with the hierarchical organization of incoming material \cite{pallier2011constituent}. This relationship may be particularly salient in Cantonese: because Chinese lacks overt inflectional morphology (such as tense or subject–verb agreement) and relies heavily on word order, aspect markers, sentence-final particles, and discourse context to signal syntactic and pragmatic relations, grammatical comprehension depends closely on tracking and updating contextual representations across successive spans. Children with stronger Cantonese grammatical proficiency may therefore integrate incoming structural cues more efficiently into the evolving narrative context, reflected in the consistent negative slope between grammar scores and semantic-update coefficients across control, attention, and sensorimotor parcels.In this context, the grammar associations are consistent with a link between knowledge of linguistic structure and the neural processing of contextual change.

Experimental work on contextual integration and updating identifies contributions from both semantic and domain-general frontal and parietal systems \cite{branzi2023context}. This organization provides a relevant framework for the parietal control and dorsal-attention findings. Somatomotor involvement additionally motivates testing a sensorimotor contribution to contextual processing. Longitudinal narrative research associates changing language and higher-order network organization with reading proficiency across development \cite{szaflarski2012longitudinal,horowitzkraus2024language}. The present analysis adds a computational variable to that systems-level perspective by linking grammar scores to signed contextual-update coefficients. It motivates experiments that vary grammatical constraints and narrative context jointly, allowing their interaction to be measured across language systems.

\subsection{Context processing links oral comprehension and literacy}

The combined findings connect shared narrative organization with individual variation in the response to context. A common model-derived trajectory predicted responses across children, while literacy and grammar covaried with regional encoding and contextual-update measures. This convergence places acquired language skills within the neural processing of the same unfolding narrative. It supports examining proficiency through the strength, sign and location of contextual associations. Each measure provides a specified view of the relationship between oral comprehension and literacy, making their convergence accessible to computational analysis.

Crucially, evaluating these brain–behavior links with locally standardized, norm-referenced instruments—HKT-P(III) \cite{ho2015hong} and HKCOLAS \cite{t2006hong}—allowed individual proficiency to be calibrated against Hong Kong age norms across primary grades 2–6, avoiding the age confounding inherent in raw task scores or indirect environmental proxies. Because these same batteries are used clinically in Hong Kong to identify specific learning disorder and developmental language disorder, the observed coupling between standardized scores and naturalistic neural measures also highlights the potential of story-listening paradigms as child-friendly, ecologically valid complements for investigating atypical language and literacy development.

\subsection{Limitations and future work}

The findings characterize individual differences in 16 assessment-complete children within a cross-sectional cohort. Larger longitudinal samples are needed to estimate the stability of the associations and distinguish skill-related variation from developmental change. The brain--behavior models considered each assessment separately, so contributions from age, general cognitive ability, residual motion, neural response reliability and correlated language skills remain to be separated. Future multivariable and longitudinal analyses can test their relative contributions and the direction of relationships between language experience and neural responses.

Encoding predictions were evaluated in held-out children listening to the same story. This design tests cross-subject correspondence for a shared stimulus; transfer to new narratives, languages and populations requires additional validation. The analysis used one model checkpoint, layer, contextual-window specification and parcellation. Comparisons across stories, models and independently specified feature sets can establish which relationships generalize and which depend on these choices.

Regional summaries selected the top 25\% of voxels using each child's imaging statistic. Held-out correlations were used both to rank voxels and to summarize performance. The reported values are therefore descriptive estimates conditional on that selection. Selection by absolute GLM $t$ likewise emphasizes voxels with larger feature-related statistics. Assessment scores were separate from voxel selection, but independent localization and voxel-fraction sensitivity analyses would clarify the stability of regional associations. BH-FDR correction controlled multiplicity within each ROI family, rather than across all assessments, features and metrics jointly. Replication and broader family definitions can assess the robustness of the resulting brain--behavior findings.

The mechanistic interpretations also require targeted testing. Model-derived entropy, surprisal and semantic update are computational proxies; coefficient signs depend on their joint design and covariance structure. Prior story-listening work reported different predominant directions for entropy and surprisal \cite{willems2016prediction}, motivating explicit comparisons of stimulus, model and regressor specifications. Haemodynamic convolution limits the temporal separation of onset- and offset-aligned events. Negative coefficients therefore do not directly identify neural inhibition, and lower encoding performance does not establish greater processing efficiency. Semantic update indexes geometric representational change, with syntactic specificity to be tested through controlled stimuli and complementary linguistic annotations. Independent functional localization, event-memory measures and multimodal recordings can evaluate the proposed cortical and subcortical interpretations. Longitudinal learning studies can further test how literacy experience changes these computational relationships.

Finally, consistent with the sociolinguistic environment of Hong Kong, all participants were native Cantonese speakers who also acquired English and Mandarin early in childhood; future work with larger samples could examine whether variation in multilingual exposure interacts with contextual narrative processing in the dominant language.

\section{Conclusion}

This study relates individual language and literacy skills to the neural processing of continuous narrative context in Cantonese-speaking children. Contextual embeddings captured responses shared across listeners, surprisal and entropy characterized complementary predictive profiles, and literacy and grammar covaried with regional encoding and contextual-update measures. Their convergence positions naturalistic context processing as a measurable link between oral language and literacy. The resulting computational perspective connects behavioral proficiency with the regional strength and direction of neural responses to context, providing a basis for investigating experience-related variation in childhood language processing.

\section*{Acknowledgment}
This work was supported by Brain Science and Brain-like Intelligence Technology - National Science and Technology Major Project (2021ZD0200500), The Hong Kong Polytechnic University Departmental Collaborative Research Fund (Project ID: P0056428), an internal grant from The Hong Kong Polytechnic University (Project ID: P0048377), The Hong Kong Polytechnic University Collaborative Research with World-leading Research Groups Fund (Project ID: P0058097) and Research Grants Council Collaborative Research Fund (Project ID: C5033-24G). The authors thank the University Research Facility in Behavioral and Systems Neuroscience (UBSN), The Hong Kong Polytechnic University, for technical and facility support. The authors thank all participants for their time and contribution to this study.

\section*{Author contributions}
D.H.: Conceptualization, Methodology, Software, Validation, Formal analysis, Data curation, Visualization, Writing - original draft. T.J.: Conceptualization, Methodology, Investigation, Validation, Data curation, Writing - original draft. N.W.: Conceptualization, Resources, Supervision, Project administration, Funding acquisition, Writing - review \& editing. W.T.S.: Conceptualization, Resources, Supervision, Project administration, Funding acquisition, Writing - review \& editing. D.H. and T.J. contributed equally to this work. All authors have read and approved the final version of the manuscript.

\section*{Data and code availability}
The data that support the findings of this study are available from the corresponding author upon reasonable request. The analysis code for this study is available at \url{https://github.com/hdy6438/contextual-language-fmri}.

\section*{Competing interests}
All authors declare no competing interests.

\bibliographystyle{unsrtnat}
\bibliography{references}
\clearpage

\section*{Supplementary Information}
\addcontentsline{toc}{section}{Supplementary Information}
\setcounter{figure}{0}
\setcounter{table}{0}
\renewcommand{\thefigure}{S\arabic{figure}}
\renewcommand{\thetable}{S\arabic{table}}
\renewcommand{\theHfigure}{supp.\arabic{figure}}
\renewcommand{\theHtable}{supp.\arabic{table}}

These supplementary displays document the reported Gemma 3 1B/Schaefer--Tian/top 25\% specification, with PC128 embeddings for encoding. They provide model parameters, all tested-family totals, all 30 group-response findings, all five assessment associations and the cohort distributions. The accompanying source tables retain all 12,808 regional tests and the 232 regional encoding-performance summaries for this specification.

\begin{table}[htbp]\centering\small
\caption{Parameters of the reported analysis specification.}\label{tab:selected_parameters}
\begin{tabularx}{\linewidth}{@{}lX@{}}\toprule Parameter & Value \\\midrule
Language model & Gemma 3 1B (\texttt{google/gemma-3-1b-pt}) \\
Checkpoint revision & \texttt{fcf18a2a879aab110ca39f8bffbccd5d49d8eb29} \\
Decoder blocks / extracted block & 26 / 13 (one-based) \\
Native hidden dimensions & 1,152 \\
Encoding representation & 128 principal components \\
Embedding variance retained & 91.97\% \\
Atlas & Schaefer 200 cortical parcels plus Tian scale-2 subcortical parcels (32) \\
Encoding ROIs / eligible text-GLM ROIs & 232 / 214 \\
Within-ROI voxel fraction & Top 25\% \\
Encoding voxel ranking & Held-out Pearson $r$ within each child and ROI \\
Text-GLM voxel ranking & Absolute first-level $t$ within each child, feature and ROI \\
Inference cohort & 16 children with complete assessment data \\
\bottomrule\end{tabularx}
\end{table}

\begin{table}[htbp]\centering\small
\caption{All tests in the reported specification and their configuration-conditional BH-FDR results.}\label{tab:counts}
\begin{tabular}{lrrrr}\toprule Analysis & Families & ROIs & Tests & $q\leq0.05$ \\\midrule
Surprisal: Group response & 1 & 214 & 214 & 14 \\
Surprisal: Assessment association & 11 & 214 & 2,354 & 0 \\
Entropy: Group response & 1 & 214 & 214 & 16 \\
Entropy: Assessment association & 11 & 214 & 2,354 & 0 \\
Semantic update: Group response & 1 & 214 & 214 & 0 \\
Semantic update: Assessment association & 11 & 214 & 2,354 & 3 \\
Encoding: partial $R^2$ & 11 & 232 & 2,552 & 1 \\
Encoding: Pearson $r$ & 11 & 232 & 2,552 & 1 \\
\midrule Total & 58 & & 12,808 & 35 \\
\bottomrule\end{tabular}
\par\smallskip\begin{minipage}{.96\linewidth}\footnotesize Each family contains all eligible ROIs for one text feature or encoding metric and one group or assessment contrast.\end{minipage}
\end{table}

\clearpage
\begingroup\fontsize{8}{10}\selectfont\setlength{\tabcolsep}{3pt}
\begin{longtable}{@{}p{57mm}rrrrr@{}}
\caption{Complete BH-FDR-positive group responses in the reported specification ($n=16$).}\label{tab:selected_group}\\
\toprule Feature and ROI & Mean $\beta$ & 95\% CI & $t$ & $p$ & $q$ \\\midrule\endfirsthead
\multicolumn{6}{l}{\textbf{Table \thetable\ (continued)}}\\\toprule Feature and ROI & Mean $\beta$ & 95\% CI & $t$ & $p$ & $q$ \\\midrule\endhead
\bottomrule\endfoot
\multicolumn{6}{l}{\textbf{Surprisal}}\\\addlinespace
pHIP-rh & $-0.0228$ & $[-0.0356, -0.0099]$ & $-3.76$ & $0.0012$ & $0.0448$ \\
mAMY-rh & $-0.0402$ & $[-0.0626, -0.0179]$ & $-3.84$ & $0.0018$ & $0.0448$ \\
THA-DP-rh & $-0.0281$ & $[-0.0444, -0.0119]$ & $-3.70$ & $0.0026$ & $0.0448$ \\
pHIP-lh & $-0.0233$ & $[-0.0357, -0.0110]$ & $-4.02$ & $0.0019$ & $0.0448$ \\
lAMY-lh & $-0.0501$ & $[-0.0724, -0.0277]$ & $-4.78$ & $4.58\times10^{-4}$ & $0.0448$ \\
mAMY-lh & $-0.0596$ & $[-0.0985, -0.0206]$ & $-3.26$ & $0.0026$ & $0.0448$ \\
NAc-core-lh & $-0.0293$ & $[-0.0448, -0.0138]$ & $-4.02$ & $0.0013$ & $0.0448$ \\
L Default\_PFC\_1 & $-0.0410$ & $[-0.0657, -0.0162]$ & $-3.53$ & $0.0029$ & $0.0448$ \\
R SalVentAttn\_TempOccPar\_1 & $-0.0382$ & $[-0.0582, -0.0181]$ & $-4.06$ & $0.0016$ & $0.0448$ \\
R Limbic\_OFC\_2 & $-0.0340$ & $[-0.0526, -0.0153]$ & $-3.88$ & $0.0011$ & $0.0448$ \\
R Cont\_PFCv\_1 & $-0.0353$ & $[-0.0555, -0.0150]$ & $-3.72$ & $0.0026$ & $0.0448$ \\
R Cont\_PFCl\_5 & $-0.0256$ & $[-0.0401, -0.0110]$ & $-3.74$ & $0.0029$ & $0.0448$ \\
R Cont\_PFCmp\_1 & $-0.0334$ & $[-0.0526, -0.0142]$ & $-3.71$ & $0.0024$ & $0.0448$ \\
R Default\_PFCv\_1 & $-0.0517$ & $[-0.0836, -0.0198]$ & $-3.45$ & $0.0020$ & $0.0448$ \\
\multicolumn{6}{l}{\textbf{Entropy}}\\\addlinespace
pHIP-rh & $0.0254$ & $[0.0146, 0.0362]$ & $5.03$ & $4.58\times10^{-4}$ & $0.0327$ \\
mAMY-rh & $0.0396$ & $[0.0188, 0.0604]$ & $4.06$ & $0.0016$ & $0.0418$ \\
pHIP-lh & $0.0262$ & $[0.0100, 0.0425]$ & $3.45$ & $0.0037$ & $0.0490$ \\
lAMY-lh & $0.0459$ & $[0.0267, 0.0651]$ & $5.10$ & $2.44\times10^{-4}$ & $0.0327$ \\
aCAU-lh & $0.0241$ & $[0.0130, 0.0352]$ & $4.62$ & $8.24\times10^{-4}$ & $0.0418$ \\
L Cont\_Cing\_2 & $0.0309$ & $[0.0125, 0.0492]$ & $3.59$ & $0.0029$ & $0.0418$ \\
L Default\_Temp\_3 & $0.0261$ & $[0.0101, 0.0422]$ & $3.46$ & $0.0029$ & $0.0418$ \\
R DorsAttn\_PrCv\_1 & $0.0226$ & $[0.0102, 0.0351]$ & $3.89$ & $0.0024$ & $0.0418$ \\
R SalVentAttn\_TempOccPar\_1 & $0.0266$ & $[0.0113, 0.0419]$ & $3.70$ & $0.0025$ & $0.0418$ \\
R SalVentAttn\_FrOperIns\_1 & $0.0248$ & $[0.0109, 0.0386]$ & $3.81$ & $0.0022$ & $0.0418$ \\
R Cont\_PFCv\_1 & $0.0275$ & $[0.0145, 0.0405]$ & $4.50$ & $4.27\times10^{-4}$ & $0.0327$ \\
R Cont\_PFCl\_3 & $0.0299$ & $[0.0133, 0.0465]$ & $3.84$ & $0.0026$ & $0.0418$ \\
R Cont\_Cing\_2 & $0.0223$ & $[0.0095, 0.0351]$ & $3.71$ & $0.0027$ & $0.0418$ \\
R Cont\_PFCmp\_1 & $0.0255$ & $[0.0126, 0.0385]$ & $4.20$ & $0.0011$ & $0.0418$ \\
R Default\_PFCv\_1 & $0.0340$ & $[0.0157, 0.0522]$ & $3.97$ & $0.0014$ & $0.0418$ \\
R Default\_PFCdPFCm\_2 & $0.0204$ & $[0.0097, 0.0310]$ & $4.08$ & $0.0017$ & $0.0418$ \\
\end{longtable}
\noindent\footnotesize Parcel names retain the Schaefer network and parcel designation or Tian anatomical designation. Coefficients are in percent-signal-change units for standardized text features. Confidence intervals are analytic $t$ intervals (15 degrees of freedom); $p$ values use two-sided exact sign flips. $q$ values use the ROI-wise BH family defined in Methods.
\endgroup

\begin{table}[htbp]\centering\small
\caption{Complete BH-FDR-positive assessment associations in the reported specification ($n=16$).}\label{tab:associations}
\begin{tabularx}{\linewidth}{@{}Xlrrr@{}}\toprule ROI and brain measure & Assessment & Effect & $p$ & $q$ \\\midrule
L SalVentAttn\_FrOperIns\_1; $R^2_{\mathrm{partial}}$ & Dictation & $-0.8542$ & $5.64\times10^{-5}$ & $0.0131$ \\
L Vis\_13; $r$ & Literacy & $-0.8174$ & $2.03\times10^{-4}$ & $0.0472$ \\
L Cont\_Par\_3; $\beta_{\mathrm{SU}}$ & Grammar & $-0.0262$ & $1.00\times10^{-4}$ & $0.0214$ \\
R SomMot\_9; $\beta_{\mathrm{SU}}$ & Grammar & $-0.0223$ & $3.00\times10^{-4}$ & $0.0285$ \\
R DorsAttn\_Post\_5; $\beta_{\mathrm{SU}}$ & Grammar & $-0.0237$ & $4.00\times10^{-4}$ & $0.0285$ \\
\bottomrule\end{tabularx}
\par\smallskip\begin{minipage}{.97\linewidth}\footnotesize Encoding effects are Spearman correlations; text-GLM effects are slopes per sample SD of the grammar score. Encoding probabilities use exact label permutations; text-GLM probabilities use 19,999 random score permutations plus the observed assignment. All $q$ values retain the original configuration-specific BH correction. The three grammar-slope 95\% CIs, in row order, are $[-0.0374, -0.0150]$, $[-0.0298, -0.0149]$, $[-0.0338, -0.0136]$.\end{minipage}
\end{table}

\begin{figure}[t]
\centering
\includegraphics[width=\linewidth]{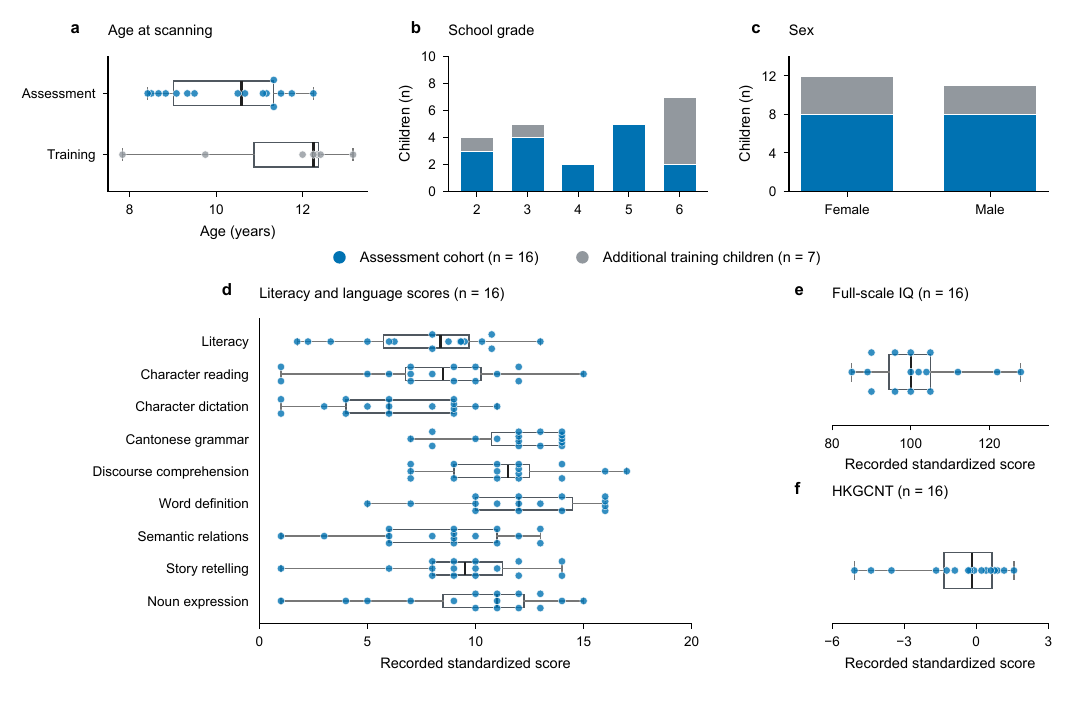}
\caption{\textbf{Cohort composition and assessment distributions.} \textbf{a}, Ages of 16 assessment-complete children and seven additional training children. \textbf{b,c}, Grade and sex counts for the 23-child imaging cohort. \textbf{d}, Nine literacy and language measures; \textbf{e,f}, full-scale IQ and HKGCNT. Dots represent children; vertical offsets separate tied scores. Boxes show medians and interquartile ranges; whiskers span the observed ranges. All 11 assessments include the same 16 children and retain their recorded standardized scales.}
\label{fig:cohort}
\end{figure}
\end{document}